\documentclass[11pt]{article}

\usepackage[preprint]{acl}

\usepackage{times}
\usepackage{latexsym}
\usepackage{booktabs}
\usepackage{multirow}
\usepackage{amsmath}

\newcommand{\ul}{\it}

\usepackage[T1]{fontenc}
\usepackage[utf8]{inputenc}

\usepackage{microtype}

\usepackage{inconsolata}

\usepackage{graphicx}
\usepackage{epstopdf}

\title{S3D: A Simple and Cost-Effective Self-Speculative \\[.5em]
Decoding Scheme for Low-Memory GPUs}

\author{Wei Zhong \hspace{5em} Manasa Bharadwaj \\
\texttt{wei.zhong@lge.com} \qquad 
\texttt{manasa.bharadwaj@lge.com} \\[0.7em]
\large LG Electronics, Toronto AI Lab
}

\begin{document}
\maketitle
\begin{abstract}
Speculative decoding (SD) has attracted a significant amount of research attention due to the substantial speedup it can achieve for LLM inference.
However, despite the high speedups they offer, speculative decoding methods often achieve optimal performance on high-end devices or with a substantial GPU memory overhead.
Given limited memory and the necessity of quantization, a high-performing model on a high-end GPU can slow down by up to 7 times. 
%
%In this work, we highlight these issues and additionally take memory into consideration when measuring speedups.
%
To this end, we propose \textbf{S}kippy \textbf{S}imultaneous \textbf{S}peculative \textbf{D}ecoding (or S3D), a cost-effective self-speculative SD method based on simultaneous multi-token decoding and mid-layer skipping.
When compared against recent effective open-source SD systems, our method has achieved one of the top performance-memory ratios while requiring minimal architecture changes and training data.
Leveraging our memory efficiency, we created a smaller yet more effective SD model based on Phi-3. It is 1.4 to 2 times faster than the quantized EAGLE model and operates in half-precision while using less VRAM.
\end{abstract}

\begin{figure}[t]
  \centering
  \includegraphics[width=1\columnwidth]{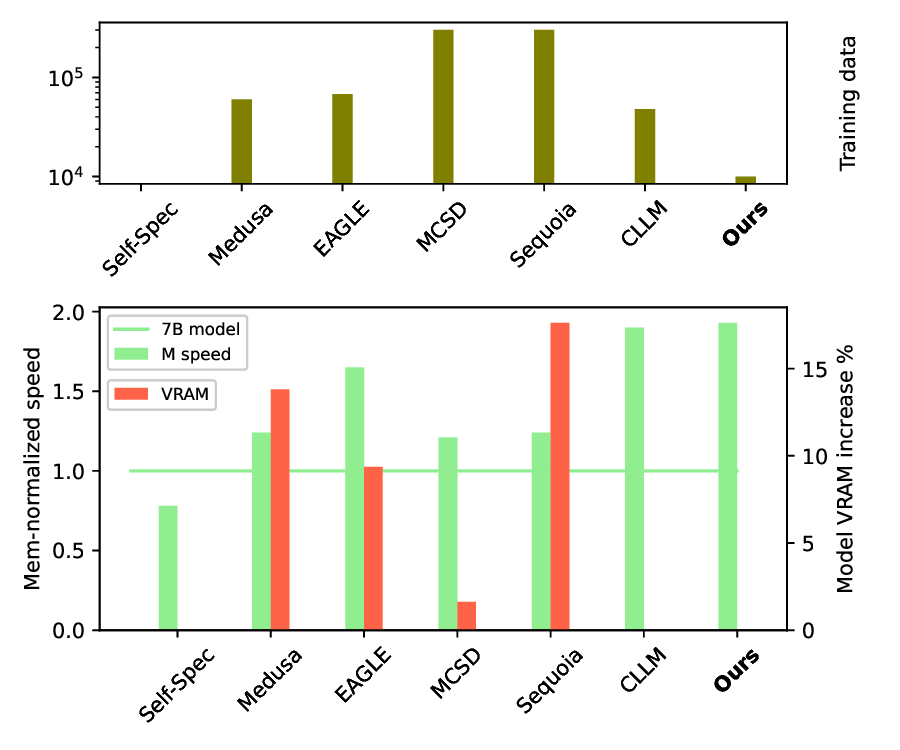}
  \caption{Training efficiency, inference efficiency per memory unit, and load-time VRAM evaluated for different models on MT-Bench. From left to right: The most recent open-source SD systems ordered by release dates. All systems use 7B target models with 8-bit quantization. Our model (S3D) stands out in both training efficiency and memory-speed trade-offs.}
  \label{fig:highlights}
\end{figure}

\section{Introduction}
Speculative decoding (SD)~\cite{stern2018blockwise,zhang2024BeyondGameSurvey,xia2024UnlockingSurvey} can accelerate LLM inference without sacrificing the quality. As a result, it is becoming one of the most common optimization techniques in LLMs.
At a high level, typical speculative decoding (SD) works by \textit{drafting} tokens at a relatively faster speed, and then \textit{verifying} the guessed tokens at the end of an iteration using a full forward pass. The speedup is based on the assumption that the accepted tokens in one forward pass during the verification step will offset the cost of the drafting steps.

However, greater speedups are not always free.
On one hand, some popular SD systems~\cite{cai2024medusa, li2024eagle, chen2024sequoia} add a considerable amount of memory, e.g., due to the extra modules or a large token trees used for drafting.
When models are deployed at scale, even a minor memory overhead can largely increase the cost of  inference, given the current high cost of using HBM VRAM in inference.
On the other hand, high-performing SD can achieve remarkable speedups using a large model~\cite{zhang2023selfspec_draftverify,zhao2024ouroboros,yang2024multicandidate} or on high-end GPUs~\cite{zhang2023selfspec_draftverify,chen2024sequoia,kou2024cllms,elhoushi2024layerskip}.
However, we notice that these speedups become limited or even start underperforming when a smaller model or low-memory device is used where quantization is necessary.
Surprisingly, the state-of-the-art open-source SD model (in speedups) may slow down by up to 7 times when applying quantization under constrained GPU memory, highlighting the significant overheads from quantization~\cite{lin2024qserve}.
In such cases, we question the cost-effectiveness of existing SD methods, even if they show greater speedups on high-end GPUs.

Although recently \citet{chen2024sequoia} designed a GPU-agnostic optimization by pre-profiling the GPU times for the draft and verify stages, their approach relies on the hard assumption of fixed acceptance rates among different levels of the draft token tree, making it less applicable to various SD methods. Additionally, the optimal trees have fewer differences for smaller models or slower GPUs, limiting their potential on low-end devices.

In this work, we introduce \textbf{S}kippy \textbf{S}imultaneous \textbf{S}peculative \textbf{D}ecoding (or S3D) to achive fast inference, low VRAM costs, and high training efficiency. Our key contributions are listed below:
\\[0.5em]
\noindent \textbf{Effective Self-speculative SD}: We propose a simple and memory-saving self-speculative decoding scheme named \textit{S3D} for low-memory GPUs. Our scheme features mid-layer skipping and simultaneous multi-token predictions, offering no added VRAM costs and high training efficiency.
Compared to \citet{zhang2023selfspec_draftverify}, S3D overcomes the limited speedups in smaller models.
\\[0.5em]
\noindent \textbf{Optimal hyper-parameters}: Instead of relying on statistical optimization, we formalize the relationship between the number of skipped layers and speedup in self-speculative decoding, as previously studied in \citet{zhang2023selfspec_draftverify} empirically. Based on our formalization, we can also verify the optimal number of token predictors, as empirically observed by \citet{gloeckle2024betterandfaster}.
%Our systematic study accelerates the hyper-parameter optimization for other model sizes as well.
\\[0.5em]
\noindent \textbf{Optimal speed-memory ratio}: Our SD method demonstrates optimal performance-memory ratios among recent open-source SD models. By exploiting the memory efficiency, we can avoid the significant quantization overheads under certain VRAM constraints and outperform the previous fastest SD method, i.e., EAGLE~\cite{li2024eagle}, under 8-bit quantization by up to 3.9x in speedups on A10G GPU.
Moreover, by switching to a smaller target model, we have created a more effective SD model based on Phi-3, which decodes 1.4 to 2 times faster than EAGLE 
 on A10G while using less VRAM.

\section{Related Work}
Early work in speculative decoding (SD) using Transformers~\cite{stern2018blockwise,sun2021earlywork_pku,xia2023speculative} focused on in-domain tasks such as translation and grammar error correction, where significant speedups are easily achieved. These methods are characterized by single-branch speculation using additional modules~\cite{stern2018blockwise} or an independent draft model~\cite{xia2023speculative}.
By using speculative sampling~\cite{deepmind2023SD}, SD can also sample tokens from target model distribution~\cite{leviathan2023SDprove}. 
In contrast, our work targets general domain tasks and focuses on greedy\footnote{Our approach can be easily extended to support sampling; we focus on greedy decoding as it is orthogonal to speculative sampling.} and non-batching decoding via \textit{simultaneous multi-token prediction}. We save memory and gain training efficiency through \textit{layer-skipping}.

\subsection{Multi-Token Predictions}
Since \citet{stern2018blockwise}, predicting the next $k$ tokens simultaneously has proven effective, but it requires adding $k$ feed-forward decoder layers from the last encoder state. \citet{cai2024medusa} popularized this idea using \textit{Medusa heads}, additionally predicting multiple token branches using tree attention~\cite{miao2024specinfer,spector2023treeattention}.
%
%The drafted tokens in a prediction tree are also dubbed a \textit{token tree}.

In the SpecDec method~\cite{xia2023speculative}, multi-token prediction is done by unmasking future tokens from multiple decoder heads attending to different encoder states, utilizing distinct attention queries for predicting different tokens. However, SpecDec requires full model fine-tuning for all layers as the decoder-only target model has not been pretrained on multi-token unmasking tasks.
More recently, \citet{apple2024stream_without_auxiliary_models} predict multiple tokens by adding streaming embeddings initialized from upper layers, with the token tree reduced by early exiting. However, they made major changes to the original architecture, and their speedups are considered limited.

Multi-token prediction can also be implemented auto-regressively~\cite{yang2024multicandidate,li2024eagle,ankner2024hydra}, which takes multiple steps to predict the next draft token conditioned on previously drafted tokens in one iteration.
To mitigate the substantial overheads incurred by multi-step drafting within a single iteration, the draft overhead should be minimal while ensuring it retains the capability to generate acceptable tokens.
In the case of EAGLE~\cite{li2024eagle},  this is achieved by efficiently concatenating the target model's high-level features with the embeddings or hidden states of the next tokens for regression via an additional layer of Transformer decoder.

Another line of work to generate multiple draft tokens in parallel is based on Jacobi iteration methods, treating auto-regressive decoding in LLM as a non-linear system of equations, or Jacobi decoding~\cite{song2021feedforward_as_eqationsolving,santilli2023JacobiDecoding}.
In practice, however, an LLM may obtain marginal speedups from Jacobi decoding as it can rarely produce an accepted token if a previous token in the trajectory is predicted incorrectly.
Lookahead decoding~\cite{fu2024lookahead} attempts to address this issue by introducing memory costs and caching n-gram tokens from previous Jacobi trajectories.
Inspired by the Consistency Model~\cite{song2023consistency_model}, CLLMs~\cite{kou2024cllms} additionally train their target models to minimize the distances between Jacobi trajectories and the fixed point, leading to faster convergence and thus greater speedups.
Compared to regular SD methods, Jacobi decoding does not have a separate draft phase.

\subsection{Layer Skipping}
Layer skipping is a type of structured pruning technique~\cite{anwar2017structured_pruning,maxwelling2018structured_pruning,danqi2022structured_pruning} that reduces a model by only using a subset of its layers.
Structured pruning is particularly intriguing for LLM optimizations due to its compatibility with GPU acceleration. This is because it enables immediate gains in memory and compute by discarding substructures entirely~\cite{van2023llm_surgeon}.

Various layer skipping schemes explored for Transformer models are discussed next. 
Early Exiting~\cite{dehghani2018universal_transformers,teerapittayanon2017branchynet} utilizes early layers and skips the rest.
LayerDrop~\cite{fan2019layerdrop} randomly drops layers during training and skips layers during inference.
\citet{lagunas2021block_pruning,van2023llm_surgeon} have identified sub-layer structures (e.g., attention heads or weight blocks) to be pruned during training.
\citet{sajjad2022poormansbert} propose symmetric dropping of complete top and bottom layers, based on the observation that middle layers of a Transformer are less critical. This observation has been validated by \citet{ma2023llmpruner,wu2024layercondensed} on larger LLMs and for KV-cache as well.
Recently, \citet{raposo2024mixture_of_depths} have trained additional parameters to dynamically skip layers.

Combining layer skipping with SD leads to an intriguing way to save memory, known as \textit{self-speculative decoding}~\cite{bae2023FREE,zhang2023selfspec_draftverify,liu2024kangaroo,elhoushi2024layerskip}, where a static or adaptive number of layers of the target model are used for drafting tokens.
These approaches mitigate the common memory overheads of SD by incorporating minimal or no extra modules for the draft stage. Specifically, they either entail no additional training~\cite{zhang2023selfspec_draftverify} or only necessitate training for learning an adaptive early exit threshold~\cite{bae2023FREE,elhoushi2024layerskip} with a potential trade-off in quality.
In \citet{elhoushi2024layerskip}, they reuse KV-cache from draft stages, reducing the computation needed for the remaining layers in the verify stage. However, their approach involves training LayerDrop~\cite{fan2019layerdrop} and requires complex early exit thresholds during inference.

A concurrent work by \citet{gloeckle2024betterandfaster} combines self-speculative decoding with multi-token predictions. However, their approach trains additional independent heads comprising entire Transformer layers, potentially adding more memory cost compared to EAGLE~\cite{li2024eagle}. In contrast, our SD scheme imposes no extra model load cost and has minimal training requirements through applying mid-layer skipping.

\section{Preliminaries}
Given a Transformer decoder-only model~\cite{radford2018gpt1} $M_p$, its generated next-token distribution $p(t_{\le i})$ given the current context tokens $t_{\le i} = t_1, t_2, ..., t_i$ can be expressed in terms of Transformer layers.
For layer $\ell=1,2,..., L$,
\begin{align}
h^{(0)}_i =& \text{Emb}(t_i) \\
h^{(\ell)}_i =& T^{(\ell)}(h^{(\ell-1)}_{\le i}, \text{Pos}_{\le i}) \\
t_{i+1} \sim& p(t_{\le i}) = \text{LM-Head}(h^{(L)}_i)
\end{align}
where $\text{Emb}$ represents the embedding transformation, and $T^{(\ell)}$ denotes the Transformer layer at level $l$, which receives the context hidden states from the previous layer $h^{(\ell-1)}_{\le i}$, associated with their position information $\text{Pos}_{\le i}$.
The $\text{LM-Head}$ maps the hidden space to a vocabulary distribution $p$ for the next token sampling.

The decoder-only language model is typically trained using the next-token prediction task, where training involves employing cross entropy loss across tokens in parallel.
Given a sample of sequential tokens $t_i, i=1,2,...,N$, the loss is
\begin{equation}
\mathcal{L} = \frac1{N-1} \sum^{N-1}_{i=1} - \log p(t_{\le i})_{ t_{i+1} }
\end{equation}

During a SD iteration, a more efficient draft model $M_q$ is often used to predict the next $\gamma$ token(s) from the target model $M_p$ through  sampling $t_{i+j+1} \sim q(t_{\le i+j})$ where $j = 0, 1, 2, ..., \gamma - 1$.
To produce tokens as if they were sampled from the target distribution, \citet{leviathan2023SDprove}  show that we can verify drafted tokens by comparing $p(t_{i+j})$ with $q(t_{i+j})$ successively,
and accept each token with a probability of 
$\min(1, \frac{p(t_{i+j})}{q(t_{i+j})})$.
Upon completion, one more last token $t_{i+\gamma+1}$ can be sampled from the target distribution $p$.
On rejection, sampling is done from a normalized distribution of $\max(0, p - q)$. 
In greedy decoding, this process is equivalent to accepting only the matched tokens produced from $p$ and $q$.

\section{S3D}

\begin{figure*}
  \centering
  \includegraphics[width=0.80\linewidth]{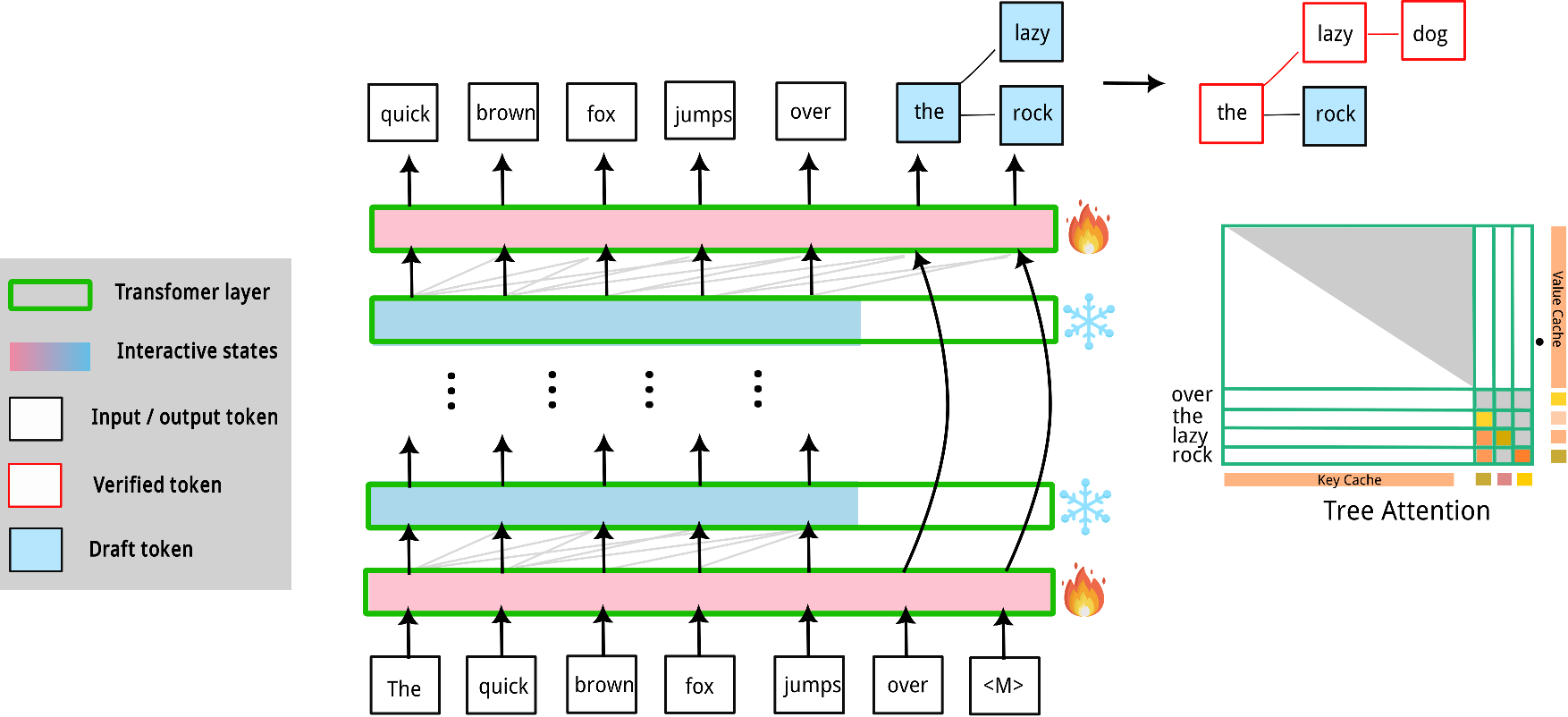}
  
  \caption{An illustration of S3D based on simultaneous predictions of the last $\gamma$ tokens ($\gamma=2$).
  A mask token \texttt{<M>} is added into vocabulary prior to training, and a partial model is trained to predict the next tokens simultaneously.
  Tree attention is adopted to verify multiple branches of predictions give top candidates of the $k$-th draft token.
  Unlike other self-speculative decoding methods based on fully-skipped layers, we only skip the middle layers on top of the draft tokens
  so that the draft model can access high-level features from top layers as well as the previous states verified by the complete target model.}
  \label{fig:architecture}
\end{figure*}

We propose a self-speculative SD scheme called \textbf{S}kippy \textbf{S}imultaneous \textbf{S}peculative \textbf{D}ecoding (or S3D).
In S3D, the draft model $M_q$ uses partial layers from target model $M_p$.

To adhere to the Transformer decoder architecture and circumvent the need for auxiliary modules, we opt to emulate the Masked Language Modeling (MLM) task commonly employed in Transformer encoder training. This involves inserting a special mask token, denoted as \texttt{<M>}, into future inputs to predict the next $\gamma$ tokens concurrently. Specifically, %draft model receives the last token, and $\gamma - 1$ mask tokens:
\begin{equation}
\label{eq:multi_token_pred}
{t}_{i+1}, ..., {t}_{i+\gamma} \sim q(t_{\le i}, \underbrace{\text{<M>}, ..., \text{<M>}}_{\gamma - 1})
\end{equation}
where the draft model $M_q$ uses all previous hidden states of the target model $h^{(\ell)}_{\le i}, \ell = 1,2,...,L$:
\begin{equation}
h^{(\ell)}_i = T^{(\ell)}(h^{(\ell-1)}_{\le i}, \text{Pos}_{\le i})
\end{equation}

Different from \citet{xia2023speculative},
the simultaneously generated tokens at $j=i+1, i+2, ..., i + \gamma$ require only propagating through lower and top layers, skipping middle $m$-th to $n$-th layers of the target model:
\begin{align}
h^{(n-1)}_j =& h^{(m)}_j \\
h^{(\ell')}_j =& T^{(\ell')}(h^{(\ell'-1)}_{\le j}, \text{Pos}_{\le j})
\end{align}where non-skipping layers $\ell'=1, 2, ..., m, n, n+1, ..., L$.
Unlike  \citet{zhang2023selfspec_draftverify}, a non-skipping layer $\ell$
is able to utilize previous states of the target model, i.e., $h^{(\ell-1)}_{\le i}$.
Furthermore, in contrast to early exiting in \citet{zhang2023selfspec_draftverify}, the current draft states from top layers are kept for decoding.
Additionally, we do not necessarily skip lower layers due to the adjustments required in lower-level representations for skipped middle layers, a notion explored similarly by \citet{ma2023llmpruner} and \citet{wu2024layercondensed}.
We will further justify this skipping scheme in Section~\ref{sect:optim_params}.
\\[.5em]
\noindent \textbf{Training}:
Our training objective is to accurately uncover masked tokens while preserving the original next-token prediction capability.
To this end, we train the draft model to decode both the next token right after $i$ and its following masked tokens.
Assume the masked tokens are located at $i+1, i+2, ..., i+\gamma - 1$, our training loss is
\begin{equation}
\label{eq:loss}
\mathcal{L}^{(\footnotesize S3D)} = \frac1{|D|} \sum_{j\in D} - \log q(t_{\le j})_{ t_{j+1} }
\end{equation}
where decoding set $D=\{i, i+1, ..., i+\gamma-1\}$.

During training, we freeze the skipped layers to preserve the target model distribution.
Instead of predicting next tokens sequentially, we  assign masked tokens randomly so that training samples can be processed in one batch, utilizing the parallelism of Transformer.   
An illustration of our modeling is shown in Figure~\ref{fig:architecture}.
\\[.5em]
\noindent \textbf{Predicting speedup}:
Given target ratio $\beta \in [0, 1]$,  which represents the ratio of target model parameters that the draft model uses during decoding, 
the acceptance rate $\alpha$ of the frist drafted token should be a function of $\beta$. 
Naturally, in the self-speculative case, $\alpha(1) = 1$ and $\alpha \to 0$ when $\beta \to 0$ .

In this work, we have hypothesized a function to estimate draft token acceptance rate as a function of model size (parameterized by $U$):
\begin{equation}
\label{eq:hypo_accept_rate}
\alpha(\beta; U) = \frac{1 - U^\beta}{1 - U}.
\end{equation}
We will show in Section~\ref{sect:optim_params} that the above function aligns well with empirical observations.

In multi-token predictions, assume the true acceptance rate at the $k$-th draft token, i.e., $\alpha_k(\beta)$, is discounted by $k$ in a discrete function (which we may readily estimate from empirical data).
Following the notation in \citet{li2024eagle}, the expected newly generated tokens $\tau$ is~\footnote{
When drafting a single next token, i.e., $\alpha_k(\beta)=\alpha_1(\beta)$, Eq.~\ref{eq_sumprod} is a capped geometric series and can be further simplified to a formula given by \citet{leviathan2023SDprove}.}
\begin{equation}
\label{eq_sumprod}
\tau(\gamma, \beta)= \sum_{n=1}^{\gamma+1} n \cdot \prod_{k=1}^{n-1} \alpha_k (\beta) \cdot z_{n}(\beta)
\end{equation}
where the shorthand notation $z_{n}(\beta) = 1 -\alpha_n(\beta)$ if $n \not= \gamma+1$ and $1$ otherwise.

When the number of drafting tokens is not significant (which commonly happens on low-end devices),
it is reasonable to assume the time cost for a full-model forward pass to be a constant $C$. 
We also assume there is a fixed overhead $H_0$ for each iteration, proportionally to $C$, i.e., $H_0 = \delta \cdot C$,
then deriving the decoding speed by taking out $C$ leaves us the \textit{Improvement Factor}~\cite{leviathan2023SDprove} for self-speculative decoding, i.e.,
\begin{equation}
\label{eq:if}
IF(\gamma, \beta) = \frac{\tau(\gamma, \beta)}{\delta + \beta + 1},
\end{equation}
assuming that the forward time for a partial model scales linearly with the number of its parameters.

Note that Eq.~\ref{eq:if} represents a different improvement factor compared to the one in auto-regressive drafting schemes~\cite{leviathan2023SDprove}, where $\gamma$ predictions are performed sequentially.

\section{Experiments}
\label{sect:experiments}

\begin{table*}[]
\caption{The cost-effectiveness comparisons on A10G GPU considering peak VRAM costs. All models are 8-bit quantized and are based on the 7B LLaMA-v2 target model except mentioned otherwise in parentheses. The largest 3 numbers in each column are highlighted in \textit{italics} or \textbf{bold}. ``Peak'' denotes the peak VRAM usage in GiB. The overall (averaged) results count for both $M$ speed and 
 the relative effectiveness metrics compared to the baseline.}
\label{tab:majortab}
\hspace{-0.8em}
\resizebox{1.03 \textwidth}{!}{
\small
\begin{tabular}{l|ccccccccccccccc}
\toprule
\multicolumn{1}{c|}{\multirow{2}{*}{\textbf{Model \textbackslash{} Metric}}} & \multicolumn{4}{c|}{\textbf{MT-Bench}} & \multicolumn{4}{c|}{\textbf{Human-Eval}} & \multicolumn{5}{c|}{\textbf{CNN-Daily}} & \multicolumn{2}{c}{\textbf{Overall}} \\ 
\multicolumn{1}{c|}{} & \multicolumn{1}{c}{\textbf{\scriptsize Peak} $\downarrow$} & \multicolumn{1}{c}{\textbf{Tok / s}} & \multicolumn{1}{c}{$M$} & \multicolumn{1}{c|}{\textbf{Score}} & \multicolumn{1}{c}{\textbf{\scriptsize Peak} $\downarrow$} & \multicolumn{1}{c}{\textbf{Tok / s}} & \multicolumn{1}{c}{$M$} & \multicolumn{1}{c|}{\textbf{Acc. \%}} & \multicolumn{1}{c}{\textbf{\scriptsize Peak} $\downarrow$} & \multicolumn{1}{c}{\textbf{Tok / s}} & \multicolumn{1}{c}{$M$} & \multicolumn{1}{c}{\textbf{R-1}} & \multicolumn{1}{c|}{\textbf{R-L}} & \multicolumn{1}{c}{$M$} & \multicolumn{1}{c}{\textbf{Eff.}} 

\\ \midrule

Baseline & {\ul 8.53} & 7.02 & 1.00 & 7.05 & {\ul 8.43} & 7.02 & 1.00 & 6.71 & {\ul 8.96} & 6.53 & 1.00 & 0.19 & 0.13 & 1.00 & 1.00 \\
Self-Spec. & 7.77 & 5.00 & 0.78 & \textbf{7.08} & 7.46 & 4.89 & 0.79 & 6.10 & 8.09 & 5.01 & 0.85 & 0.19 & \textbf{0.14} & 0.81 & 0.97 \\
\scriptsize Medusa (Vicuna) & {\ul 9.09} & 9.27 & 1.24 & 4.98 & {\ul 8.94} & 10.69 & 1.44 & 7.93 & {\ul 9.36} & 7.62 & 1.12 & \textbf{0.24} & \textbf{0.14} & 1.26 & 1.03 \\
EAGLE & {\ul 9.58} & \textbf{13.03} & 1.65 & 6.98 & {\ul 9.45} & \textbf{15.22} & 1.93 & 5.49 & {\ul 9.86} & \textbf{12.43} & \textbf{1.73} & 0.19 & 0.13 & 1.77 & 0.94 \\
MCSD & 7.76 & 7.72 & 1.21 & 6.79 & 7.40 & 7.77 & 1.26 & \textbf{9.76} & 8.10 & 6.74 & 1.14 & 0.20 & \textbf{0.14} & 1.21 & \textbf{1.16} \\
Sequoia & 8.44 & 8.64 & 1.24 & 6.46 & 8.35 & 9.01 & 1.30 & 3.05 & 8.57 & 7.99 & 1.28 & 0.18 & 0.12 & 1.27 & 0.77 \\
Ouroboros & 7.95 & 5.47 & 0.84 & \textbf{7.08} & 7.61 & 5.83 & 0.92 & \textbf{8.54} & 8.30 & 4.91 & 0.81 & 0.18 & 0.13 & 0.86 & \textbf{1.09} \\
CLLM & 7.51 & 11.75 & \textbf{1.90} & 5.31 & 7.37 & \textbf{16.29} & \textbf{2.66} & 3.66 & 7.53 & 8.06 & 1.47 & 0.20 & \textbf{0.14} & \textbf{2.01} & 0.79 \\[0.5em]
\bf Ours &&&&&&&&&&&&&&& \\
S3D & 7.79 & \textbf{12.39} & \textbf{1.93} & 5.68 & 7.60 & 13.85 & \textbf{2.19} & 6.71 & 8.80 & \textbf{9.58} & \textbf{1.49} & \textbf{0.28} & \textbf{0.19} & \textbf{1.87} & \textbf{1.09} \\
\scriptsize S3D (Phi-3, fp16) & 8.14 & \textbf{25.31} & \textbf{3.78} & \textbf{7.04} & 7.92 & \textbf{28.13} & \textbf{4.27} & \textbf{20.12} & 8.87 & \textbf{17.35} & \textbf{2.69} & \textbf{0.25} & \textbf{0.18} & \textbf{3.58} & \textbf{1.77}\\

\bottomrule
\end{tabular}
}
\end{table*}

\subsection{Experimental Setup}
\noindent \textbf{Datasets} We consider datasets commonly used in SD evaluations, including MT-Bench~\cite{zheng2023mt_bench} for multi-turn conversation, Human-Eval~\cite{chen2021human_eval} for coding, and CNN-Daily~\cite{hermann2015cnn_daily} for summarization.
For CNN-Daily, we only use the $1,000$ samples similar to \citet{zhang2023selfspec_draftverify}, while the complete datasets are used for the others.
For MT-Bench, we use Gemini Pro~\cite{team2023gemini} for effectiveness evaluation.
We report accuracy scores for Human-Eval and Rouge-1 and Rouge-L scores~\cite{lin2004rouge} for CNN-Daily.
\\[.5em]
\noindent \textbf{Baselines} 
A fair comparison is conducted by running all systems on identical samples and hardware using a uniform evaluation framework.
To this end, our model is compared to open-source SD systems including: Self-Spec~\cite{zhang2023selfspec_draftverify},
Medusa~\cite{cai2024medusa},
EAGLE~\cite{li2024eagle}, 
MCSD~\cite{yang2024multicandidate}, Sequoia~\cite{chen2024sequoia},
Ouroboros~\cite{zhao2024ouroboros},
and CLLMs~\cite{kou2024cllms}.
Self-Spec is a training-free self-speculative method that predicts the next single token via layer skipping.
Medusa is a popular SD method that adds parallel decoder heads to predict multiple next tokens.
EAGLE concatenates target model late-layer hidden states with the last token embedding to predict the next 5 tokens auto-regressively via an additional Transfomer decoder layer.
And recent work MCSD, Sequoia, and Ouroboros generate draft tokens through a separate draft model.
In particular, Sequoia constructs an optimal draft token tree from profiling the underlying hardware.
In these three systems, we adopt the 68M JackFram LLaMA~\cite{miao2024specinfer} as the draft model, which is also the default and most efficient option for their LLaMA target models.
Lastly, CLLMs is considered as the latest development in the direction of Jacobi or Lookahead decoding~\cite{santilli2023JacobiDecoding,fu2024lookahead}.
\\[.5em]
\noindent \textbf{Implementation details} 
All implementations use eager-mode decoding based on Huggingface Transformers~\cite{wolf2020huggingfaces} and we adopt the native BNB 8-bit quantization~\cite{dettmers2022bnb8bit} for quantized models.
We configure each system with greedy decoding and keep the other SD configurations default for different systems. 
A non-speculative implementation is used as the common baseline to calculate speedups and relative memory costs as it shares the same instructional LLaMA-v2~\cite{touvron2023llama} backbone for most of our evaluated systems.

We mostly consider the 7B target model size. In exception to this, we also train a 3.8B Phi-3 Mini~\cite{abdin2024phi3} target model and a 13B LLaMA-v2 target model to demonstrate the generalization ability of our method.
Unless specified otherwise, our S3D implementations use the optimal hyper-parameters suggested by Section~\ref{sect:optim_params}.
\\[.5em]
\noindent \textbf{Cost effectiveness metric} 
We propose a memory-normalized speed metric $M$, which divides a speedup by the relative memory usage compared to the baseline model:
\begin{equation}
 M = \frac{v_1}{v_0} / \frac{m_1}{m_0} = \frac{v_1}{m_1} / \frac{v_0}{m_0}   
\end{equation}
where \( v_0 \) and \( m_0 \) are the generation speed and memory consumption of the baseline model, and \( v_1 \) and \( m_1 \) are the generation speed and memory consumption of the evaluating model.
This metric quantifies the generation speedup per memory unit, ensuring a fair and memory-aware comparison for target models of the same size.
\\[.5em]
\noindent \textbf{Training}
Similar to Medusa, EAGLE, CLLMs, et al.\cite{cai2024medusa,li2024eagle,kou2024cllms}, we train our models on the  ShareGPT dataset. 
All training is conducted using bf16 and FlashAttention-2~\cite{dao2023flashattention2} with a batch size of 64 for one epoch on A10G GPUs.
Please refer to Appendix~\ref{sect:detailed_train_config}
for detailed training descriptions.

\subsection{Main Results}
Initially, we discuss the cost-effectiveness of our model. In Table~\ref{tab:majortab} and \ref{tab:majortab2}, we observe that S3D models and CLLM exhibit the highest overall $M$ speeds among the evaluated systems, which remains consistent across different GPUs.

Importantly, models producing high speedups are not necessarily the most cost-effective, as seen in the case of Medusa and EAGLE, where the cost of extra draft module(s) must be considered.
We also observed discrepancies in the effectiveness scores of models when using the original target model for token verification, likely due to implementation issues or numerical errors. However, our model generally maintains baseline effectiveness and achieves the highest overall effectiveness among systems requiring target model training (i.e., Self-Spec., CLLM, and ours).
By using layer adapters like LoRA~\cite{hu2021lora}, we can easily enable lossless decoding at the expense of  efficiency costs (see analysis in Appendix~\ref{sect:lora}). 

Interestingly, the vanilla self-speculative decoding method, i.e., Self-Spec, underperforms the baseline in terms of speed for the 7B target model. This highlights the limitation of naive self-speculative decoding ($\gamma=1$) in smaller models, where the partial draft model becomes further constrained and unable to propose good draft tokens.
In contrast, we alleviate this issue by allowing the draft model to attend to previous target model states and training the model to predict multiple tokens ($\gamma > 1$), thereby enhancing the effectiveness of smaller self-speculative models.

\begin{figure}[t]
  \centering
  \includegraphics[width=0.9\columnwidth]{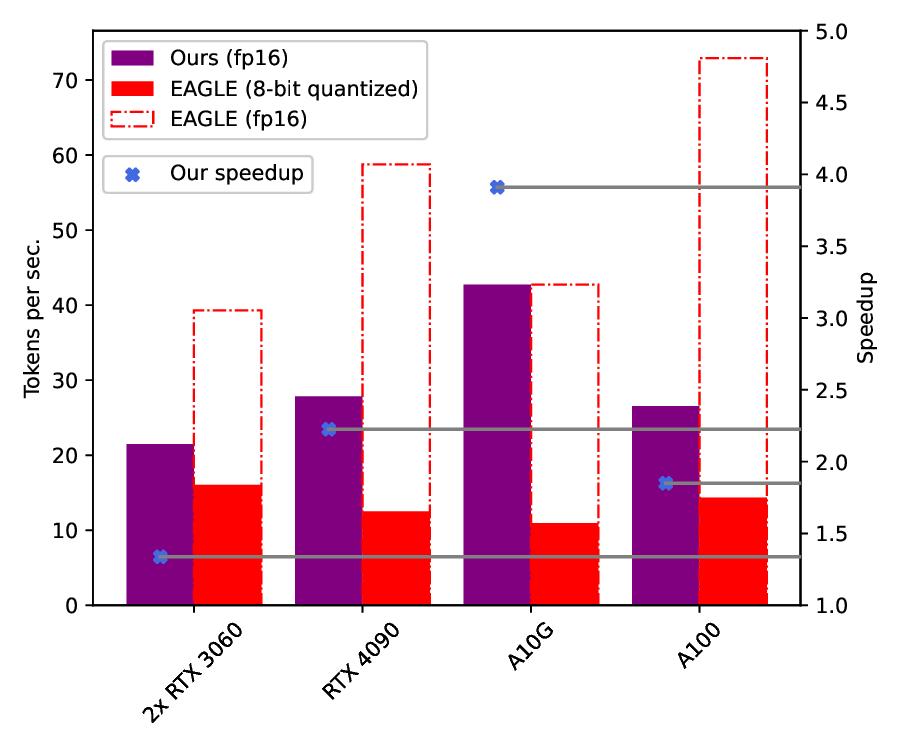}
  \caption{Speed comparison between ours (S3D) and EAGLE on different GPU devices (MT-Bench samples, 7B LLaMA target model). The dashed bars represents the full speed potentials of the EAGLE model without memory restrictions.  However, when constrained with a VRAM limit of 16 GiB, the quantized EAGLE model (indicated by red bars) suffers from severe speed degradation, highlighting the significant overheads associated with quantization.}
  \label{fig:diff_gpus}
\end{figure}

Admittedly, the LLaMA-based S3D model ties closely to EAGLE and underperforms CLLM in overall $M$ speed, primarily due to the high speedups and optimal memory efficiency achieved by EAGLE and CLLM, respectively.
However, as shown in Table~\ref{tab:majortab} and \ref{tab:majortab2},
we are able to exploit our memory efficiency and outperform EAGLE in both efficiency and effectiveness while using less amount of VRAM by switching to a non-quantized Phi-3 target model.
\footnote{
We conducted ablations in Appendix~\ref{sect:phi3ablation} to understand the significant improvements in Phi-3 S3D.
For a larger 13B model, our cost-effectiveness can be maintained as well (see Appendix Table~\ref{tab:majortab3}).}
Even without switching to a different target model, we demonstrate in Figure~\ref{fig:diff_gpus} that our LLaMA-based S3D model can operate in half-precision within a VRAM limit of 16 GiB, and outperform EAGLE by up to 3.9 times when EAGLE needs to be quantized. 
This underscores the critical importance of memory efficiency.

On the other hand, we find that the training objectives of CLLM may encourage repeating patterns in its outputs, leading to degraded effectiveness scores, as seen in Table~\ref{tab:majortab} and additional case studies in Appendix~\ref{sect:cllm_casestudy}.
In contrast, our model can preserve effectiveness scores more robustly while achieving the optimal speed-memory ratios.

\subsection{Optimal Hyper-Parameters}
\label{sect:optim_params}
We first study the optimal layer skipping choices. 
To this end, we have empirically explored three different schemes by skipping a fixed number of original layers of LLaMA 7B target model:
(1) Skipping asymmetrically from the middle, including early existing and using only late layers while skipping all early layers.
(2) Skipping symmetric layers from the middle, i.e., layer 5 $\to$ 26 or skipping the middle 20 layers.
(3) Alternate evenly between skipping and non-skipping layers.

\begin{figure}
    \hspace{-1.2em}
    \includegraphics[width=1.05\columnwidth]{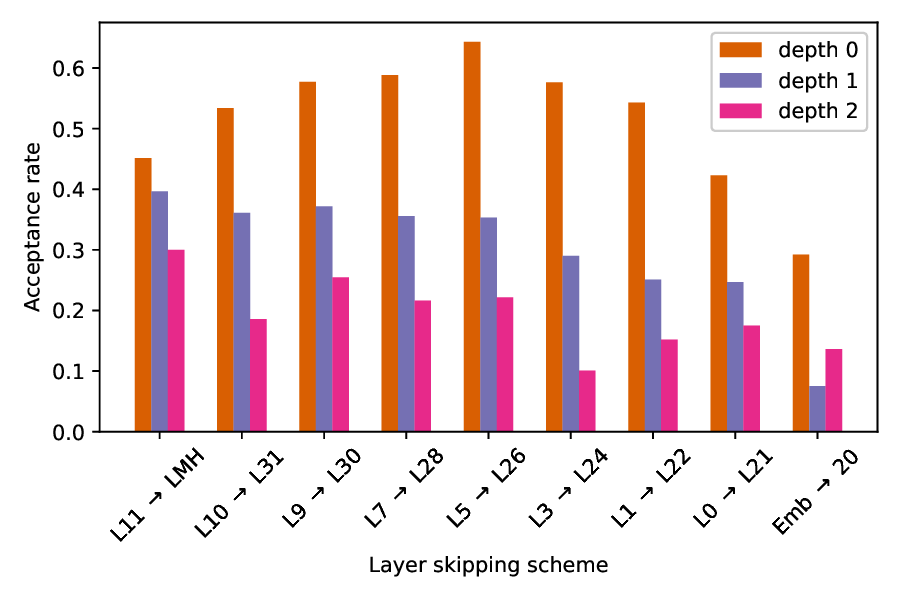}
  \caption{The overall acceptance rates and individual acceptance rates at different drafting depths (w/ only a single branch of future tokens). 
  L, LMH, and Emb stand for regular layer, LM heads, and the embedding layer, respectively.
  Skipping the middle layers symmetrically has shown better acceptance rates in general.
  Note that we distinguish embedding layer and lm\_head here although in practice they may have tied weights.
  }
  \label{fig:acceptance_rates}
\end{figure}

Looking at Figure~\ref{fig:acceptance_rates} (and Appendix Figure~\ref{fig:diff_layer} for training efficiencies), skipping symmetrical layers from the middle performs best, achieving higher overall acceptance rates and optimal training efficiency.
In contrast, skipping from layer 11 to the top (LM-head) layer and skipping from the bottom (embedding) layer to layer 20 have the worst performance, highlighting the importance of both early and late layers.
Lastly, skipping symmetric middle layers or alternating every 3 layers has similar training efficiencies.

\begin{figure}[t!]
  \centering
  \hspace{-0.6em}
  \includegraphics[width=0.81\columnwidth]{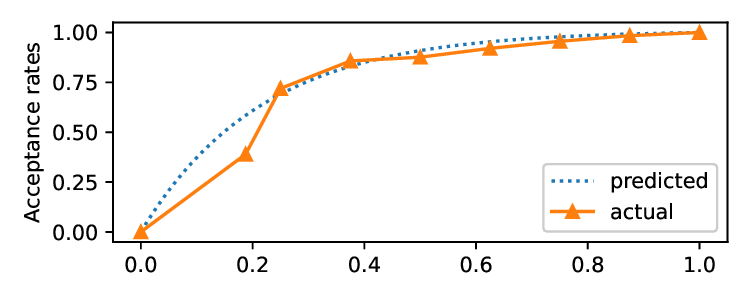}
  \hspace{0.5em}
  \includegraphics[width=0.8\columnwidth]{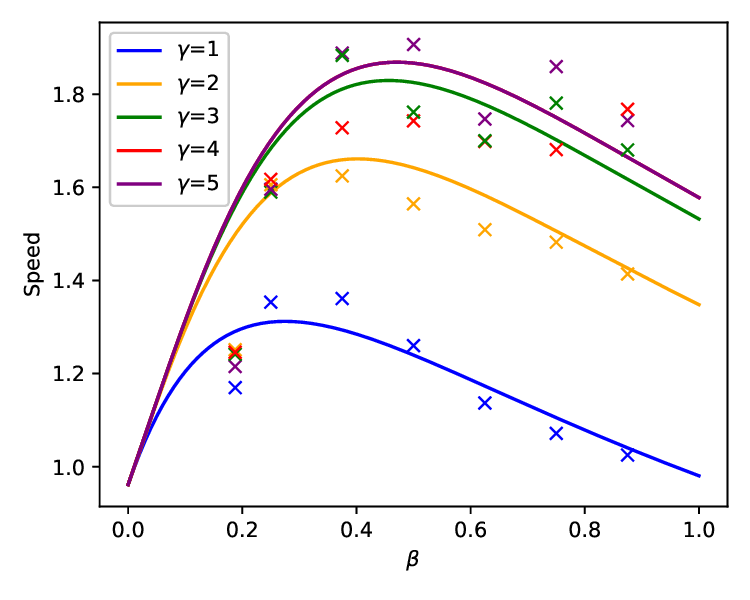}
  \caption{
  \textbf{Upper}:
  The predicted (in dashes) and sampled acceptance rates (interpolated orange dots) of various draft model sizes ($\beta$).
  \textbf{Lower}:
  The predicted (in curves) and sampled (in dots) speeds of different draft model sizes and different number of guesses ($\gamma$).
  Our prediction curves justify the optimality of using around half the number of parameters and $\gamma=4$, as observed individually and respectively in \citet{zhang2023selfspec_draftverify} and \citet{gloeckle2024betterandfaster}.
  }
  \label{fig:fig_d_and_beta}
\end{figure}

To answer what is the optimal number of layers and what is the best number of tokens to be predicted in parallel, we train different number of layers ($\beta$) skipped symmetrically from the middle layer, each trained model is evaluated for different $\gamma$ values up to $5$.
We run different models on MT-Bench for 50 samples,
and linearly interpolate the acceptance rate discount function $\alpha_k(\beta)$ (detailed in Appendix~\ref{sect:predict_formula_details}).
As summarized in Figure~\ref{fig:fig_d_and_beta},
our proposed formula for predicting self-speculative acceptance rates in Eq.~\ref{eq:hypo_accept_rate} mostly matches with the empirical results except for the lowest $\beta$ value, and this outlier may be explained by the less predictability in training a small partial model ($< 1.5$B).
Addtionally, the speedup formula in Eq.~\ref{eq:if} successfully predicts both the trend and the sweet spot in speedups.
Higher $\gamma$ values align less with the prediction because the acceptance rates for far-future tokens have higher uncertainty and variance as reflected by Figure~\ref{fig:acceptance_rates}.

In addition to the findings from \citet{zhang2023selfspec_draftverify,gloeckle2024betterandfaster}, we have unified multi-token predictions with layer skipping.
Our prediction in Eq.~\ref{eq:if} has also justified their findings that the optimal speed is achieved by skipping around half of the layers ($\beta \to 0.5$) and the optimal $\gamma$ is $4$ (as shown in Figure \ref{fig:fig_d_and_beta}, a higher $\gamma$ results in an almost diminished speedup, offset by fewer accepted tokens). 

\subsection{Training Efficiency}
In addition to its cost-effectiveness, S3D also demonstrates greater training efficiency compared to other effective SD models (see Figure~\ref{fig:highlights}).
So we hypothesize that the self-speculative decoding method used in S3D inherently lowers training costs, as the training task leverages the existing model weights.
In light of this, we compare and train both the S3D and EAGLE models using $20,000$ data rows (the original EAGLE was trained on $68,000$ data rows).

\begin{figure}[t!]
  \hspace{1em}
  \includegraphics[width=0.91\linewidth]{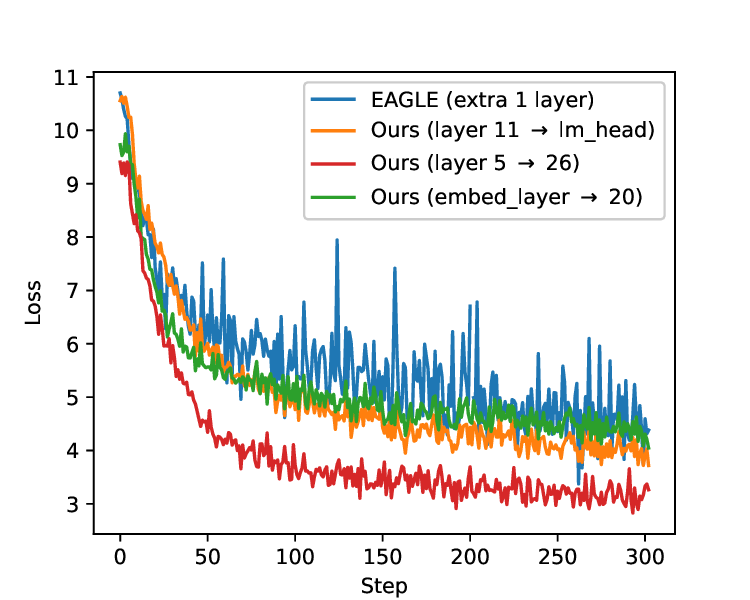} 
  \caption {Training loss comparisons between EAGLE's  classification loss~\cite{li2024eagle} and our (S3D) training loss in Eq.~\ref{eq:loss}. EAGLE requires training an extra layer of Transformer with additional linear mappings.}
  \label{fig:eagle_train_loss_compare}
\end{figure}

As shown in Figure~\ref{fig:eagle_train_loss_compare}, S3D consistently exhibits lower training losses, even when considering more layers and including far-future tokens, which are generally challenging to predict.
In the worst-performing case, where early exiting or early layer skipping is used, our loss values remain mostly below those of EAGLE.
Additionally, S3D training loss shows less variance and converges more steadily.

\section{Conclusion}
We have proposed S3D, a self-speculative SD method based on simultaneous multi-token predictions and mid-layer skipping.
S3D demonstrates one of the best cost-effectiveness among recent open SD systems, while also exhibiting high training efficiency and maintaining the effectiveness of the original model.
We have also verified the optimal hyper-parameters for our proposed method in a principled manner, without requiring any black-box optimizations beforehand.
By leveraging memory efficiency, S3D can avoid quantization and surpass the speed of quantized EAGLE when a 16 GiB VRAM limit is imposed. Additionally, S3D, based on the smaller Phi-3 target model, decodes 1.4 to 2 times faster than quantized EAGLE on A10G, with reduced VRAM usage and better effectiveness.

\raggedbottom

\section*{Acknowledgments}
We extend our gratitude to Touqir Sajed for his suggestions and brainstorming throughout the preparation of this paper. Additionally, we thank Kevin Ferreira, Yipeng (Penny) Ji, and Paria Nejat for their support in arranging computational resources.

\bibliography{custom}

\clearpage

\appendix

\section*{Appendix}

\section{Training Configurations}
\label{sect:detailed_train_config}
We train 7B models with a per-device batch size of 2, a max sequence length of 2048, and \texttt{acc\_grad}=8 with 4 GPUs. 
For 13B models, we use a per-device batch size of 1, a max sequence length of 1024, and \texttt{acc\_grad}=8 with 8 GPUs parallelized in Zero-3 using DeepSpeed with parameter offloading~\cite{rajbhandari2020deepspeed_zero}.

We uniformly apply a learning rate of $3 \cdot 10^{-5}$ with 50 linear warm-up steps and a fused AdamW \cite{loshchilov2019adamW} kernel from PyTorch for optimization.
During our multi-token prediction training, we sample $20,000$ data rows and mask out 15\% tokens randomly.

\section{S3D using LoRA}
\label{sect:lora}
Using an S3D model fine-tuned with adapters can easily guarantee the same output as the original target model, thus maintaining the original model quality.
With this intention, we have trained S3D models using LoRA~\cite{hu2021lora} with the same amount of data as full fine-tuning.

\begin{figure*}
    \centering
    \includegraphics[width=0.95\textwidth]{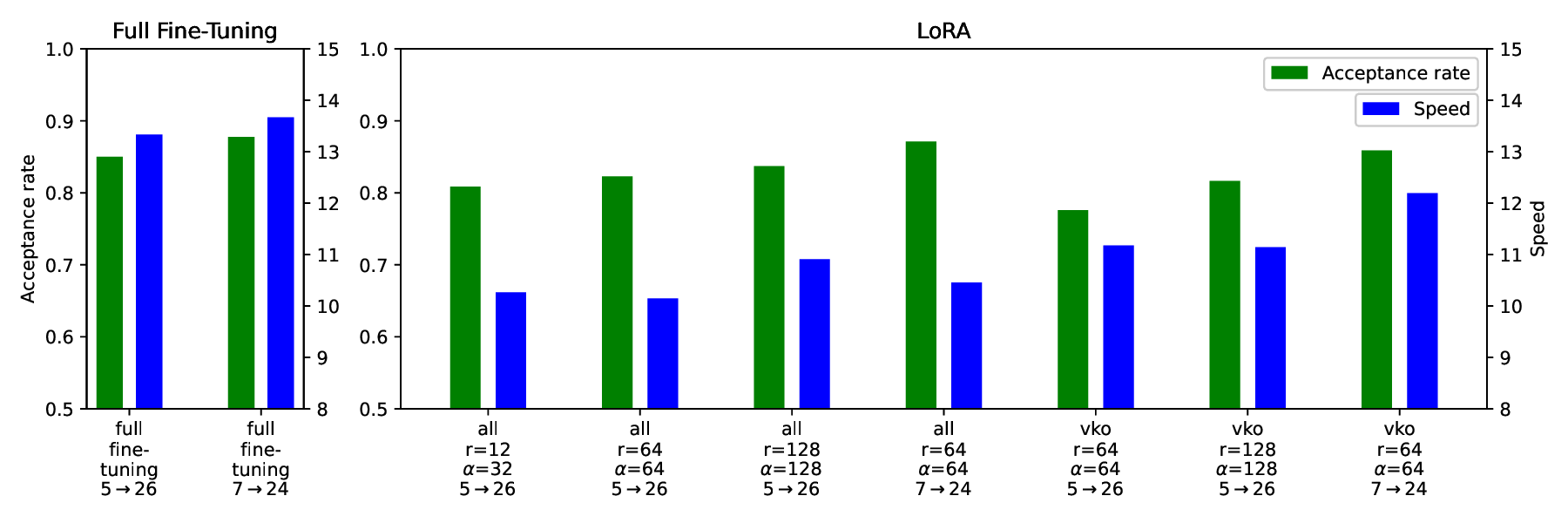}
  \caption{S3D full fine-tuning compared with using LoRA adapters (MT-Bench). LoRA settings include attaching adapters to all linear layers (all) or only attaching to value, key, and output projection layers (vko).
  Given the same layer skipping scheme, vko and low-rank LoRA have lower inference overheads but achieve lower acceptance rates (at depth-0). Overall, LoRA does not offer similar speeds compared to full fine-tuning, although it reliably maintains the original model output.
}
  \label{fig:lora}
\end{figure*}

We apply LoRA to the optimal layer skipping schemes ($5\to 26$ and $7\to 24$) and vary matrix ranks $r$ from $12$ to $128$, and $\alpha$ from $32$ to $128$.
However, as shown by Figure~\ref{fig:lora}, the speed of a LoRA-based S3D model is notably penalized by the overheads from the attached linear adapters, and the acceptance rate underperforms compared to that of a fully fine-tuned model.
As a result, we observe inferior speed when using S3D with LoRA adapters.
We have also experimented AdaLoRA~\cite{zhang2023adalora} but have observed similar negative results for end-to-end speeds. 

Nevertheless, we have seen that the BitDelta method~\cite{liu2024bitdelta} successfully compressed fine-tuned adapter weights by more than 10x. However, BitDelta requires large training resources and needs to load three models simultaneously.
Despite this, we still see great potential to achieve similar efficiency to a fully fine-tuned S3D model with guaranteed original model outputs using adapters, but we leave this to future work.

\section{Phi-3 S3D Ablation Study}
\label{sect:phi3ablation}
To comprehend the high cost-effectiveness of the Phi-3 based S3D model, we conducted an ablation study, dissecting the contributions of the target model, quantization, and the S3D method.

From Table~\ref{tab:majortab4}, it is observed that avoiding quantization generally only enhances speed, while the significant improvements in cost-effectiveness stem from two main factors: 1) transitioning to Phi-3 Mini as the target model, and 2) implementing the S3D method.
However, the latter is crucial for surpassing the state-of-the-art SD in both speedups and cost-effectiveness under limited memory.

\section{Issues of CLLMs}
\label{sect:cllm_casestudy}
Similar to our method, CLLMs~\cite{kou2024cllms} incurs no additional memory costs to the original model. It iterates and verifies multiple tokens in parallel using Jacobi iterations with random initial guess tokens. Consequently, in Table~\ref{tab:majortab} and \ref{tab:majortab2}, CLLM has achieved higher speed-memory ratios comparable to ours.

However, we have discovered that the CLLM model we have evaluated is prone to generating repetitive patterns, presumably due to its training objectives, which encourage the target model to shorten the Jacobi trajectories before reaching the fixed point. While this can accelerate inference convergence and mitigate the issue of relatively slow speedups from vanilla Jacobi decoding, we have observed relatively low effectiveness scores from the officially trained 7B model using ShareGPT data (shown in Table~\ref{tab:majortab}).
In contrast, S3D preserves most of the effectiveness of the original model.

In Figure~\ref{fig:cllm_case_study}, we present a comparison between two example outputs of CLLM. The one with repetitive patterns can generate tokens 2x faster than its counterpart, raising questions about the speed optimization of CLLM in terms of preserving the effectiveness of the original model.

\section{Prediction Formula Details}
\label{sect:predict_formula_details}
\begin{figure*}
\centering
\includegraphics[width=0.8\linewidth]{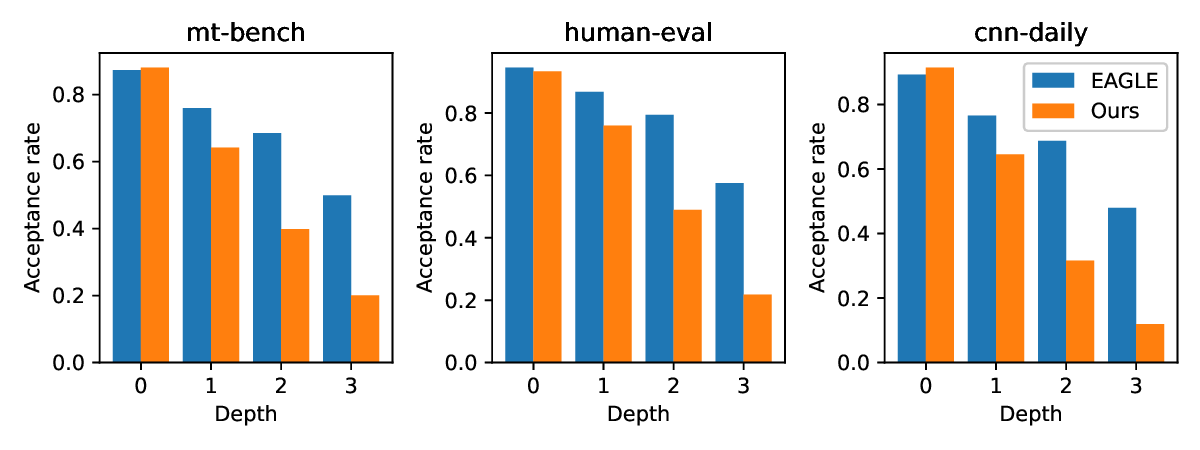}
\caption{Acceptance rates comparison at different draft token tree depths, compared between ours (S3D) and the state-of-the-art open-source model EAGLE in different datasets. Because we predict tokens simultaneously, our acceptance rates drop more at future positions. However, we achieve similar acceptance rates at the draft root and are able to outperform EAGLE cost effectively while using much less training data.}
\label{fig:acceptance_rates_cmp}
\end{figure*}

Our prediction for both acceptance rates and speeds shown in Fig~\ref{fig:fig_d_and_beta} can be captured by only a few hyper-parameters.
For Eq.~\ref{eq:hypo_accept_rate}, we use $U = 0.01$.
For SD iteration speed (Eq.~\ref{eq:if}), we set $\delta = 0.04$.
For the discount function of the $k$-th token acceptance rate, we use a linear interpolation of real data from experiments shown in Figure~\ref{fig:acceptance_rates_cmp}. Specifically,
\begin{align}
    \sigma_k(\beta)=(1-\frac {k-1}4) \cdot \sigma(\beta)
\end{align}
for $k \le 5$.

\begin{table*}
\caption{
The cost-effectiveness comparisons on a low-end RTX 3060 GPU considering peak VRAM costs. All models are 8-bit quantized and are based on the 7B LLaMA-v2 target model except mentioned otherwise in parentheses. The largest three numbers in each column are highlighted in \textit{italics} or \textbf{bold}. ``Peak'' denotes the peak VRAM usage in GiB. 
Effectiveness metrics are omitted here as they mirror those in Table~\ref{tab:majortab}.
}

\label{tab:majortab2}
\centering
\hspace{-1.2em}
\resizebox{0.9 \textwidth}{!}{

\begin{tabular}{l|cccccccccc}
\toprule
\multicolumn{1}{c|}{\multirow{2}{*}{\textbf{Model \textbackslash{} Metric}}} & \multicolumn{3}{c|}{\textbf{MT-Bench}} & \multicolumn{3}{c|}{\textbf{Human-Eval}} & \multicolumn{3}{c|}{\textbf{CNN-Daily}} & \multicolumn{1}{c}{\textbf{Overall}} \\ 
\multicolumn{1}{c|}{} & \multicolumn{1}{l}{\textbf{Peak} $\downarrow$} & \multicolumn{1}{l}{\textbf{Tok / s}} & \multicolumn{1}{l|}{\textbf{$M$}} & \multicolumn{1}{l}{\textbf{Peak} $\downarrow$} & \multicolumn{1}{l}{\textbf{Tok / s}} & \multicolumn{1}{l|}{\textbf{$M$}} & \multicolumn{1}{l}{\textbf{Peak} $\downarrow$} & \multicolumn{1}{l}{\textbf{Tok / s}} & \multicolumn{1}{l|}{\textbf{$M$}} & \multicolumn{1}{c}{\textbf{$M$}} \\
\midrule
Baseline & 8.42 & 8.11 & 1.00 & {\ul 8.35} & 7.82 & 1.00 & {\ul 8.90} & 7.42 & 1.00 & 1.00 \\
Self-Spec. & 7.81 & 5.71 & 0.76 & 7.46 & 5.46 & 0.78 & 8.08 & 5.68 & 0.84 & 0.79 \\
Medusa (Vicuna) & {\ul 9.09} & 10.32 & 1.18 & {\ul 8.94} & 11.97 & 1.43 & {\ul 9.36} & 7.99 & 1.02 & 1.21 \\
EAGLE & {\ul 9.59} & \textbf{15.18} & 1.64 & {\ul 9.45} & \textbf{17.39} & 1.97 & {\ul 9.86} & \textbf{13.82} & \textbf{1.68} & 1.76 \\
MCSD & 7.78 & 8.51 & 1.14 & 7.42 & 8.46 & 1.22 & 8.07 & 7.01 & 1.04 & 1.13 \\
Sequoia & {\ul 8.44} & 10.29 & 1.27 & {\ul 8.35} & 10.48 & 1.34 & 8.57 & 9.01 & 1.26 & 1.29 \\
Ouroboros & 7.94 & 6.28 & 0.82 & 7.60 & 6.79 & 0.95 & 8.23 & 5.62 & 0.82 & 0.86 \\
CLLM & 7.47 & \textbf{14.14} & \textbf{1.97} & 7.39 & \textbf{19.19} & \textbf{2.77} & 7.53 & 9.28 & \textbf{1.48} & \textbf{2.07} \\[0.5em]
\bf Ours &&&&&&&&&& \\
S3D & 7.81 & 13.99 & \textbf{1.86} & 7.58 & 14.99 & \textbf{2.11} & 8.72 & \textbf{10.41} & 1.43 & \textbf{1.80} \\
S3D (Phi-3, fp16) & 8.15 & \textbf{29.88} & \textbf{3.23} & 7.92 & \textbf{33.70} & \textbf{3.18} & 8.81 & \textbf{19.49} & \textbf{2.65} & \textbf{3.23} \\
\bottomrule
\end{tabular}
}
\end{table*}

\begin{table*}[]
\caption{The cost-effectiveness comparison considering peak VRAM costs for selected models using 13B 8-bit quantized LLaMA as the target model (on A10G).
The largest number in each column are highlighted in \textit{italics} or \textbf{bold}. ``Peak'' denotes the peak VRAM usage in GiB. The overall (averaged) results count for both $M$ speed and 
 the relative effectiveness metrics compared to the baseline.}
\label{tab:majortab3}
\centering
\hspace{-1.2em}
\resizebox{1 \textwidth}{!}{

\begin{tabular}{l|cccc|cccc|ccccc|cc}
\toprule
\multicolumn{1}{c|}{\multirow{2}{*}{\textbf{Model \textbackslash{} Metric}}} & \multicolumn{4}{c|}{\textbf{MT-Bench}} & \multicolumn{4}{c|}{\textbf{Human-Eval}} & \multicolumn{5}{c|}{\textbf{CNN-Daily}} & \multicolumn{2}{c}{\textbf{Overall}} \\
\multicolumn{1}{c|}{} & \textbf{Peak} $\downarrow$ & \textbf{Tok / s} & \textbf{$M$ } & \textbf{Score} & \textbf{Peak $\downarrow$} & \textbf{Tok / s} & \textbf{$M$} & \textbf{Acc. \%} & \textbf{Peak} $\downarrow$ & \textbf{Tok / s} & \textbf{$M$} & \textbf{R-1} & \textbf{R-L} & \textbf{$M$} & \textbf{Eff.} \\
\midrule
Baseline (13B) & 15.39 & 5.60 & 1.00 & 7.36 & 15.03 & 5.51 & 1.00 & 6.10 & 15.88 & 5.09 & 1.00 & 0.21 & 0.15 & 1.00 & 1.00 \\
EAGLE (13B) & {\ul 16.96} & 10.67 & 1.73 & \textbf{7.40} & {\ul 16.82} & \textbf{12.58} & 2.04 & 4.27 & {\ul 17.29} & \textbf{9.82} & \textbf{1.77} & 0.21 & 0.15 & 1.85 & 0.90 \\
S3D (13B) & 14.39 & \textbf{10.70} & \textbf{2.04} & 6.18 & 14.08 & 12.23 & \textbf{2.37} & \textbf{6.71} & 15.80 & 7.87 & 1.55 & \textbf{0.25} & \textbf{0.17} & \textbf{1.99} & \textbf{1.04} \\
\bottomrule
\end{tabular}

}
\end{table*}

\begin{figure*}
\centering

\fbox{\begin{minipage}{\dimexpr\textwidth-2cm}\itshape
USER: tell me a few interesting facts about the sun and the moon.                    
\newline
\noindent\rule{\textwidth}{1pt}

ASSISTANT: 1. The sun is approximately 4.6 million miles away from Earth,while the moon is about 238,900 miles away.
2. The sun is about 864,000 miles in diameter, while the moon is only about 2,159 miles in diameter.
3. The sun is made up of hydrogen and helium, while the moon is primarily composed of silicon and iron.
4. The sun is a spherical object, while the moon has a slightly irregularshape due to its formation process.
5. The sun takes approximately 24 days to complete one rotation, while the moon takes approximately 27.3 days to complete one rotation around Earth.
6. The sun is approximately4.6 billion years old, while the moon is estimated to be around4.5 billion years old.
7. The sun's surface temperature is approximately 5,500°F (3,000°C), while the moon's surface temperature ranges from -273°F (-170°C) during the night to 260°F (127°C) during the day.
8. The sun is a star, while the moon is a natural satellite of Earth.
9. The sun is the source of all energy on Earth,while the moon has no intrinsic energy source.
10. The sun is the largest object in the solar system, while the moon is the largest natural satellite in the solar system.
\end{minipage}

}

\vspace{3em}

\fbox{\begin{minipage}{\dimexpr\textwidth-2cm}{\itshape
USER: tell me a few interesting facts about the sun and the moon.                    
\newline
\noindent\rule{\textwidth}{1pt}

ASSISTANT: 1. The sun is a gigantic ball of plasma, while the moonis a rocky body.                                     
2. The sun is approximately 4.6 billion years old, while the moon is approximately 4.5 billion yearsold.                                                                                                                                      
3. The sun is approximately 109 times larger thanthe moon, and the moon is approximately 300 times smaller than thesun.                                                                                                                       
4. The sun is approximately 100 times more massivethan the moon.                                                       
5. The sun is approximately 93 million miles(150 million kilometers) away from Earth, while the moon isapproximately 238,900 miles (384,400 kilometers) away.
6. The sun is approximately 10,000 degrees Fahrenheit (5,538 degrees Celsius) at its surface, while the moon's surface temperature rangesfrom -273 to -233 degrees Fahrenheit (120 to -150 degrees Celsius).
7. Thesun is approximately 864,000 miles (1,392,000 kilometers) in diameter, while the moon isapproximately 2160 miles (3474 kilometers) indiameter.
8. The sun is approximately 109 times more luminous than the moon.                                                     
9. The sun is approximately 10,000 times more massive than the moon.                                                   
10. Thesun is approximately 10,000 times more luminous thanthe moon.         
11. The sun is approximately 10,000 times more massive than the moon.
12. The sun isapproximately 10,000 times more luminous than the moon.
13. The sun is approximately 10,000times more massive than the moon.
14. The sun is approximately 10,000 times more luminous than the moon.
15. The sun is approximately 10,000 times moremassive than the moon.
16. The sun is approximately 10,000 times more luminous than the moon.
17. The sun is approximately 10,000 times more massive thanthe moon.
18. The sun is approximately 10,000 times more luminous than the moon.
19. Thesun is approximately 10,000 times more massive than the moon.
20. The sun is approximately 10,000times more luminous than the moon.
}
\textbf{\small ... (repetitive pattern continues)}
\end{minipage}}

\caption{Examples of CLLM outputs (on a RTX 3060 GPU with 8-bit quantization). The implementation is non-deterministic due to potential numerical errors. As a result, we are able to show two different inputs from the same prompt. \textbf{Upper}: A good example which has a speed of 10.85 tokens per second. \textbf{Lower}: A bad example which shows repetitive patterns at the end, having a 2x higher speed of 21.98 tokens per second.}
\label{fig:cllm_case_study}
\end{figure*}

\begin{table*}[]
\caption{Ablations on the cost-effectiveness of S3D models based on Phi-3 Mini (on A10G), considering peak VRAM costs.
The largest two numbers in each column are highlighted in \textit{italics} or \textbf{bold}.
``Peak'' denotes the peak VRAM usage in GiB. 
Although using non-quantized fp16 generally improves model speeds, it does not neccessarily add cost-effectiveness due to the additional memory overheads.
}
\label{tab:majortab4}
\centering
\small
\hspace{-1.2em}
\resizebox{1 \textwidth}{!}{

\begin{tabular}{l|ccc|ccc|ccc|c}
\toprule
\multicolumn{1}{c|}{\multirow{2}{*}{\textbf{Models \textbackslash{}  Metrics}}} & \multicolumn{3}{c|}{\textbf{MT-Bench}} & \multicolumn{3}{c|}{\textbf{Human-Eval}} & \multicolumn{3}{c}{\textbf{CNN-Daily}} & \multicolumn{1}{|c}{\textbf{Overall}} \\
\multicolumn{1}{c|}{} & \textbf{Peak} $\downarrow$ & \textbf{Tok / s} & \textbf{$M$} & \textbf{Peak} $\downarrow$ & \textbf{Tok / s} & \textbf{$M$} & \textbf{Peak} $\downarrow$ & \textbf{Tok / s} & \textbf{$M$} & \textbf{$M$} \\
\midrule
LLaMA (8-bit) & {\ul 8.53} & 7.02 & 1.00 & {\ul 8.43} & 7.02 & 1.00 & {\ul 8.96} & 6.53 & 1.00 & 1.00 \\
Phi-3 (8-bit) & 4.64 & 8.91 & 2.33 & 3.78 & 8.82 & 2.80 & 5.49 & 7.99 & 2.00 & 2.38 \\
Phi-3 (fp16) & 7.99 & \textbf{16.43} & 2.50 & 8.22 & 15.97 & 2.33 & 8.80 & \textbf{13.59} & 2.12 & 2.32 \\
Phi-3 (S3D, 8-bit) & 4.79 & 14.39 & \textbf{3.65} & 5.11 & \textbf{26.71} & \textbf{6.28} & 5.53 & 11.11 & \textbf{2.76} & \textbf{4.23} \\
Phi-3 (S3D, fp16) & {\ul 8.19} & \textbf{26.00} & \textbf{3.86} & {\ul 8.56} & \textbf{48.23} & \textbf{6.77} & {\ul 8.88} & \textbf{16.95} & \textbf{2.62} & \textbf{4.42} \\
\bottomrule
\end{tabular}

}
\end{table*}

%%%%%%%%%%%%%%%%% Last page %%%%%%%%%%%%%%%%%%%

\begin{figure*}
    \centering
    \includegraphics[width=0.7\textwidth]{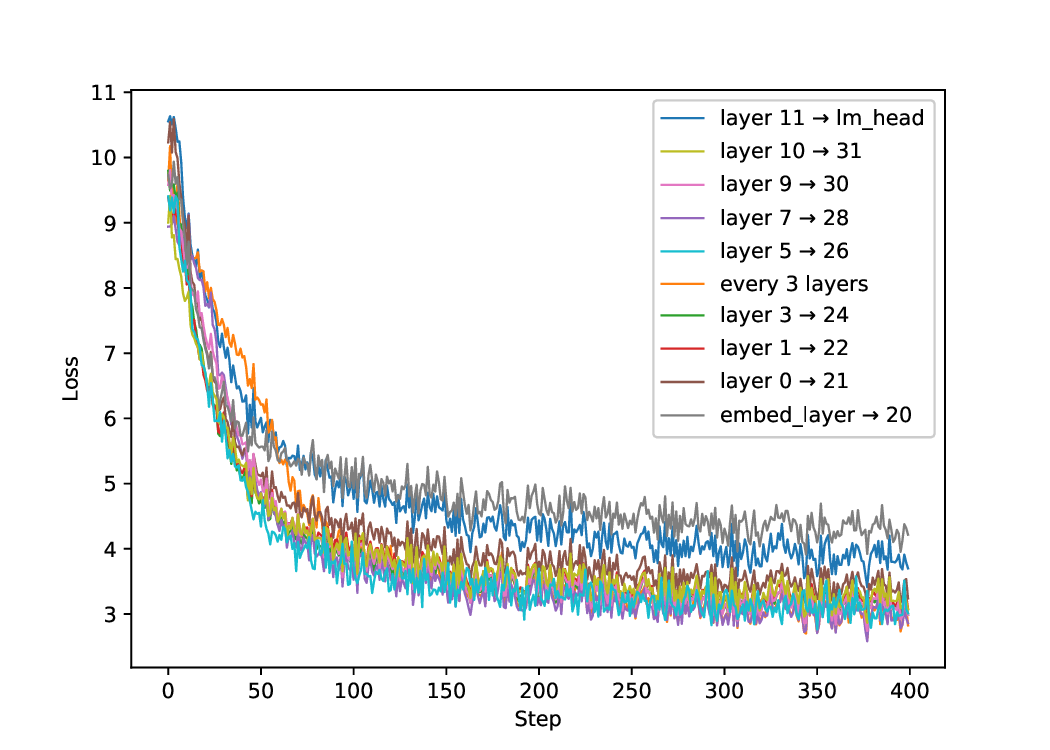}
  \caption{The training loss when fine-tuning different skipping schemes. We alter the skipped layers while keeping the total number of used layers (12 layers or $\beta=0.375$) unchanged.}
  \label{fig:diff_layer}
\end{figure*}

\end{document}